\documentclass[conference,review]{IEEEtran}
\IEEEoverridecommandlockouts
\usepackage{cite}
\usepackage{amsmath,amssymb,amsfonts}
\usepackage{algorithmic}
\usepackage{times}  
\usepackage{helvet}  
\usepackage{courier} 
\usepackage{pifont}
\usepackage{bbm}
\usepackage{bm}
\usepackage{mathrsfs}
\usepackage{graphicx}
\usepackage{textcomp}
\usepackage{xcolor}
\usepackage{booktabs}
\usepackage{multirow}
\usepackage{fancyhdr}
\usepackage{verbatim}
\usepackage{subfigure}

\def\BibTeX{{\rm B\kern-.05em{\sc i\kern-.025em b}\kern-.08em
    T\kern-.1667em\lower.7ex\hbox{E}\kern-.125emX}}
    
\begin{document}

\pagestyle{plain}

\title{ALT: An Automatic System for Long Tail Scenario Modeling
}

\author{
\IEEEauthorblockN{Ya-Lin Zhang, 
                Jun Zhou$\dagger$, 
                Yankun Ren,
                Yue Zhang,
                Xinxing Yang,
                Meng Li,
                Qitao Shi,
                Longfei Li\thanks{$\dagger$: Corresponding Author.}
      }
\IEEEauthorblockA{
	\textit{Ant Group, Hangzhou, China}\\
	\{lyn.zyl, 
            jun.zhoujun, 
            yankun.ryk, 
            yue.zy, 
            xinxing.yangxx, 
            lm168260, 
            qitao.sqt, 
            longyao.llf\}@antgroup.com}
}

\maketitle

\begin{abstract}
    In this paper, we consider the problem of \textit{long tail scenario modeling with budget limitation}, i.e., insufficient human resources for model training stage and limited time and computing resources for model inference stage. This problem is widely encountered in various applications, yet has received deficient attention so far. We present an automatic system named ALT to deal with this problem. Several efforts are taken to improve the algorithms used in our system, such as employing various automatic machine learning related techniques, adopting the meta learning philosophy, and proposing an essential budget-limited neural architecture search method, etc. Moreover, to build the system, many optimizations are performed from a systematic perspective, and essential modules are armed, making the system more feasible and efficient. We perform abundant experiments to validate the effectiveness of our system and demonstrate the usefulness of the critical modules in our system. Moreover, online results are provided, which fully verified the efficacy of our system.

\end{abstract}


\section{Introduction}
\label{introduction}
In recent years, with the prosperity and development of machine learning techniques and their increasing application to real-world tasks, unprecedented promotion has been obtained among a wide spectrum of fields, such as advertising~\cite{sun2015causal,DBLP:conf/kdd/GuoJZZYXPN0XYJZ21,DBLP:conf/sigir/LacerdaCGFZR06}, forecasting~\cite{yu2021graph,DBLP:conf/nips/ZhuLJZCZZ21,DBLP:conf/pakdd/ChenLZLQJZ19}, recommendation systems~\cite{davidson2010youtube,DBLP:conf/cikm/XuFZCZRWZ0020,DBLP:conf/icde/ChenYWBSK22,DBLP:conf/sigir/0001OM22}, image recognition~\cite{DBLP:conf/cvpr/HeZRS16,DBLP:conf/iccv/ZhengFML17,DBLP:conf/cvpr/ZophVSL18,DBLP:conf/cvpr/XieTGWYL20}, and fraud detection~\cite{zhang2019distributed,DBLP:conf/kdd/WangLZHHLZB21,DBLP:conf/www/ZhangLZLZ18,DBLP:conf/dasfaa/LiZSYCLZ21,li2022adaptiverule}, etc.
Note that typically overmuch efforts have been taken to those mainstream and important scenarios due to their considerable implication, while the attention to \textit{long tail scenarios} remains insufficient, not to mention further requirements such as the \textit{budget limitation} along this problem.

We address that the demand for \textit{long tail scenario modeling with budget limitation} is frequently encountered in our daily life.
For example, in the credit field, the platform may need to provide risk control services for numerous involved banks, while initiates may get involved periodically, with the need of enabling risk control systems for them.
Note that, from our practical experiences, specializing a model for each bank becomes necessary since that those preliminary solutions such as a unified model for all participants always lead to unsatisfactory performance.
Typically, insufficient human resources are available for the model training process, making it impossible to manually fine-tune the specific models for each of the involved banks.
At the same time, the models are required to be lightweight, since limited time and computing resources can be ensured for the model inference stage.~\footnote{A simple illustration can be found in Fig.~\ref{Fig:risk control illstration}, and more details of this example can be found in~\ref{problem statement}.}
A similar example can be found in the advertising field, where different advertisers may be attendant and new advertisers will join at any time, raising the requirements for the platform to provide effective support for these advertisers.
These examples roughly show us the prevalence of this problem.

In this work, we summarize the above problem as \textit{long tail scenario modeling with budget limitation}.
To provide a clearer definition, we broadly regard each of the involved scenarios, including both the initially existing scenarios and the subsequently emerging scenarios, as a \textit{long tail scenario}, for which a specific model is needed.
And the \textit{budget limitation} mainly refers to two aspects in this problem. 
That is the \textit{human resources limitation} for the \textit{model training stage}, and the \textit{time and computing resources limitation} for the \textit{model inference (prediction) stage}, which should be considered and solved while constructing the models.
Note that the time and computing resources are not strictly constrained during the model training stage.

Several \textit{challenges} should be emphasized while handling this problem. 
To highlight a few: 
1) \textit{Insufficient samples} may be accessible for some scenarios, which means that building an effective model from these few samples becomes challenging.
2) \textit{Budgeted time and computing resources} may be available for the \textit{inference stage}, raising the need of \textit{lightweight} yet effective models.
3) \textit{Emerging scenarios} may be frequently arise, while \textit{limited human resources} can be assigned for the \textit{model training process} of these scenarios.

To overcome the aforementioned challenges, we take efforts from several perspectives.
Roughly speaking: 
1) We leverage the samples from the related scenarios to enhance the performance of each scenario, following the philosophy of meta learning.
2) We take account of various automatic machine learning related techniques, such as hyperparameter optimization and neural architecture search, to improve the performance of the system. Specifically, to obtain effective yet lightweight models, we propose a budget-limited neural architecture search method, along with employing the distillation strategy.
3) We build an automatic system with the mentioned techniques so that the human resources can be liberated, and the whole pipeline can be automatically performed to obtain the required model whenever a new scenario arises.

In this paper, we present \textbf{ALT}, an \textbf{A}utomatic system for \textbf{L}ong \textbf{T}ail scenario modeling with budget limitation in industrial settings.
To enhance the whole framework, several automatic machine learning related techniques are introduced and improved in our system.
Namely, to get the utmost out of the data of all related scenarios, a \textit{scenario agnostic heavy model} is constructed and maintained, whose effectiveness is of great importance for the subsequent procedure and the whole system.
Note that to obtain better performance for the heavy model, we build the initial model by means of both parameter tuning for pre-designed architecture and automatic neural architecture search (NAS) techniques, and select the better one as the initial heavy model.
When it comes to a specific scenario, following the philosophy of meta learning, a copy of the scenario agnostic heavy model is first generated and then fine-tuned with the data of this scenario, arriving at a \textit{scenario specific heavy model}. 
Then to meet the requirement of budget limitation, a \textit{scenario specific light model} is built by a budget-limited NAS method proposed in this paper, along with distilling the knowledge of the \textit{scenario specific heavy model} to the constructed light model so that better performance can be achieved with high efficiency.

Moreover, a thorough system is established and presented in this paper, with several necessary modules, such as feature factory, data preparation, and model serving module, etc. 
And also, some critical optimizations from the engineering perspective are performed in our system.
Essentially, the system can automatically execute when a new scenario emerges, with the guarantee of effectiveness and efficiency, and this system can be widely used for similar tasks in an industry standard as mentioned above.

The contributions of this work can be roughly summarized as follows:
\begin{itemize}
    \item To the best of our knowledge, we are the first to present a system for the problem of modeling for long tail scenarios with budget limitation, which is frequently encountered and needed in various fields, yet barely addressed. 
    \item Several efforts are taken to improve the algorithms, including generating and maintaining a model agnostic heavy model to leverage the information of previous knowledge, refining the heavy model to obtain scenario specific heavy model, generating the scenario specific light model by proposing a budget-limited NAS method, along with distilling the knowledge of the heavy model to achieve better performance and efficiency.
    \item Several efforts are taken to enhance the whole system, including offering the necessary modules, such as feature factory, data preparation, and model serving, and performing some critical optimizations from the engineering perspective. 
    \item Abundant experimental results are reported and thorough analyses are presented to demonstrate the effectiveness of the proposed system. Online application is also presented for better demonstration.
\end{itemize}

In the rest, After discussing the background in section~\ref{background}, we introduce the proposed method and the whole system in section~\ref{method} and \ref{system}. Then the experimental results and further analyses are provided in section~\ref{experiment}. Finally, we discuss the related works and conclude this work.

\section{Background}
\label{background}
In this section, to give a clearer explanation of the studied problem, we first present the problem statement.
After that, we summarize the main challenges of this problem and present our main ideas for solving the challenges.

\subsection{Problem Statement}
\label{problem statement}

In this paper, we consider the problem of \textit{long tail scenario modeling with budget limitation}.
Basically,
the \textit{long tail scenario modeling problem} refers to the settings in which numerous participants may be already involved, while new participants will emerge periodically. Each of these participants is with the need of building a specific model. 
Herein, we regard each of the involved participants, including both the initially existing ones and the subsequently emerging ones, as a long tail scenario.
Moreover, the \textit{budget limitation} in our setting mainly refers to the \textit{human resources limitation} for the \textit{model training stage}, and the \textit{time and computing resources limitation} for the \textit{model inference stage}.

\begin{figure}[htbp]
    \centering
    \subfigure[]{
        \label{Fig:risk control illstration-a}
        \includegraphics[width=0.15\textwidth]{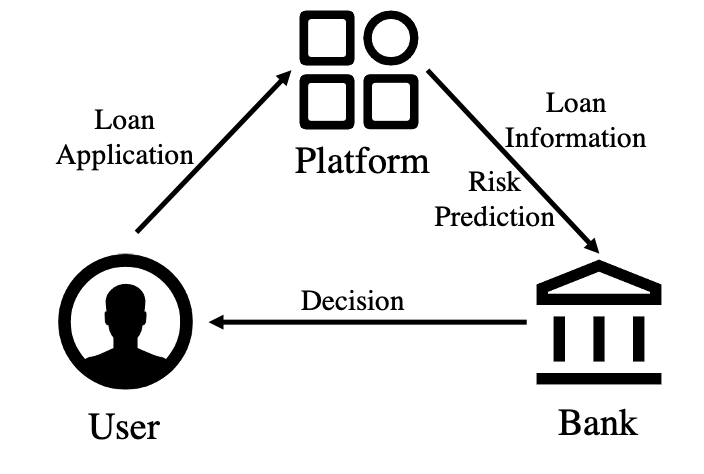}
    }
    \subfigure[]{
        \label{Fig:risk control illstration-b}
        \includegraphics[width=0.30\textwidth]{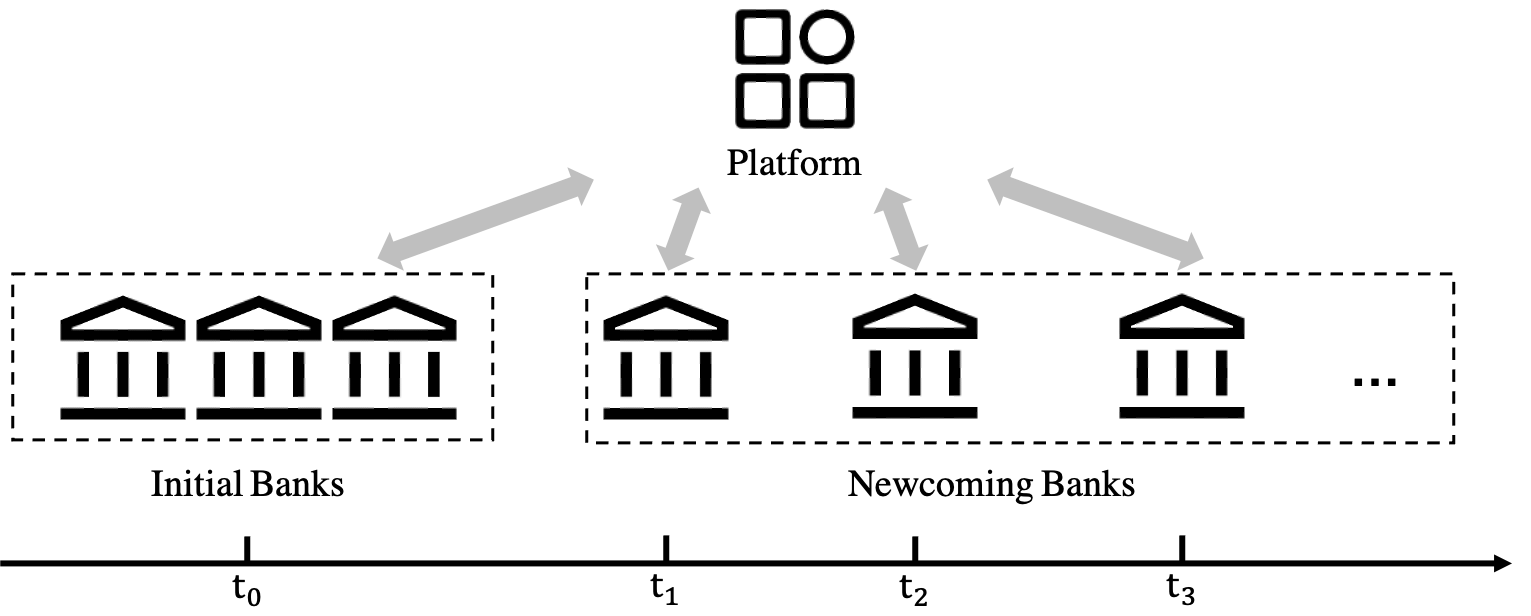}
    }
    \caption{A simple illustration of the long tail scenario modeling problem for the risk control application.}
    \label{Fig:risk control illstration}
\end{figure}

To further explain the setting and the motivation, an illustration of risk control application is shown in Fig.~\ref{Fig:risk control illstration}.
Let's assume that we hold a platform that provides risk control services for different banks, and we need to predict the probability of default for a loan application for the involved banks.
As shown in Fig.~\ref{Fig:risk control illstration-a}, for the risk control application, the model will be invoked when a user initiates a loan application through the platform and predicts the risk of this application.
Then the loan application information and the risk information will be delivered to the signatory bank of this user, and the bank will make a final decision for this loan application, i.e., whether to approve the loan application for this user.
As for the platform, the risk control service may be required from all of the involved banks, which include both the initial ones and the new-coming ones as shown in Fig.~\ref{Fig:risk control illstration-b}. 

As a preliminary solution, we may consider building a unified system for all of the participants with all available data collected from them.
However, one reality is that the requirement of more specific model is frequently posed by some participants if we follow the trivial and convenient solution, since the unified model is not specified (and thus incongruous) for these participants, and the performance is somehow unsatisfactory for them.
Moreover, newcoming banks may emerge at any time, also with the need for a specific model.
These situations show us the necessity to build a specific model for each participant.
We regard the model construction for each participant as a specific \textit{long tail scenario} and summarize this problem as a \textit{long tail scenario modeling problem}.

Note that from a more realistic perspective, further restrictions may be existing when we come to real industrial tasks.
As a nonnegligible feature, a \textit{budget limitation} is always accompanied by these tasks, bringing challenges from two main aspects.
For one thing, numerous scenarios may already exist, and new scenarios can appear at any time, while the investment of human resources is unlikely to increase accordingly. 
As an example, hundreds or even thousands of participants may get involved in the above-mentioned risk control application.
This means that \textit{human resources limitation} is encountered in the \textit{model training stage} and an automatic pipeline is essentially needed.
For another, the \textit{time and computing resources} are always limited for each scenario in the \textit{model inference stage}, so that more scenarios can be supported and the time delay of online service is acceptable for a better user experience. 
As an example, the response time may be preferred in milliseconds in our practical applications.
This means that a \textit{lightweight} model is required for these tasks.
Together with the \textit{budget limitation} situation, we come into the problem of \textit{long tail scenario modeling with budget limitation}.

Formally speaking, the goal of the system is to construct an efficient and lightweight model $f_n$ (with the parameters $\theta_n$) for each long tail scenario $\mathcal{S}_n$ of an application field with distribution $p(\mathcal{S})$, and the complexity of the model $f_n$, denoting as $\Omega(f_n)$ or $\Omega(\theta_n)$, satisfies the budget limitation (e.g., the FLOPs of the model should be under a certain threshold).
Typically, in the beginning, we are with the data collected from $N_0$ scenarios, i.e., $\mathcal{D}_n=\{\bm{x}^n_i,y^n_i\}_{i=1}^{M_n}$, $n=1,2,\cdots,N_0$, in which $\bm{x}^n_i$ represents the features of the $i$-th sample in scenario $n$, and $y^n_i$ is its corresponding label, and $M_n$ is the sample size of the $n$-th scenario.
The specific model $f_n$ for each scenario $\mathcal{S}_n$ should be then constructed for serving.
Besides the seen scenarios at the beginning, new related scenarios may emerge from time to time. We denote the emerging scenario as $\mathcal{S}_u$, and the corresponding data as $\mathcal{D}_u=\{\bm{x}^u_i,y^u_i\}_{i=1}^{M_u}$. Then a specific model $f_u$ is required, which should be automatically generated by the proposed system, with the complexity limitation satisfied.

\subsection{Main Challenges}
Several challenges should be addressed and resolved for this problem.
Concretely, 
\begin{itemize}
    \item \textit{Insufficient samples} may be available for some of the scenarios, and building an effective model by only using these samples is pretty challenging. In industrial practices, for example, we frequently encounter scenarios with less than ten thousands of samples, and for some credit scenarios, the sample size may be even smaller. 
    \item \textit{Limited time and computing resources} can be ensured for each scenario in the model inference stage. Note that numerous scenarios may be involved, and the request should be responded to in a timely manner, which means that the model inference stage should be efficient. Thus, the deployed serving models are preferred to be \textit{lightweight} yet effective.
    \item \textit{Emerging scenarios} may appear one after another, while the investment of human resources cannot increase accordingly. Thus, it is impossible to manually fine-tune each model for the involved scenarios. An automatic pipeline that can provide effective models should be armed for this difficulty.
\end{itemize}

To alleviate the above challenges, the main efforts in this work can be concluded from some aspects.
Firstly, the samples from the related scenarios are adequately leveraged, with the philosophy of meta learning and transfer learning.
Secondly, we propose a budget-limited neural architecture search method and employ the distillation strategy, so that the deployed models are more likely to be lightweight and effective.
Thirdly, we build an automatic system, with various automatic machine learning related techniques used, and essential systematic modules are also armed.
The whole system can be automatically invoked when a new scenario arises and will arrive at a lightweight and effective model.

\section{Method}
\label{method}

\subsection{Overview}
In this section, we present the method from an algorithm perspective.
Generally, the algorithm design across the whole system is with two crucial goals, i.e., obtaining superior performance for each scenario, and satisfying the budget requirements.
Apart from maintaining a unified model for all scenarios, another trivial solution is to build a specific model when encountering a new scenario, using the scenario specific samples.
However, there are often not sufficient samples for building an effective model in some scenarios.

In this work, following the philosophy of knowledge sharing, we take full advantage of the data from all scenarios. 
Specifically, a \textit{scenario agnostic heavy model} is first constructed with all initial samples from all scenarios and then updated along with the modeling of each specific scenario.
Note that this model can be adequately heavy if necessary, as long as it can achieve better performance.
This model will extract the knowledge of all scenarios and benefit the model construction procedure of each specific scenario.
Namely, when it comes to a specific scenario, the \textit{scenario agnostic heavy model} is first copied and then fine-tuned with the samples of this scenario, so that a \textit{scenario specific heavy model} can be obtained efficiently, with the ability to provide considerable performance for this specific scenario.
After that, a \textit{scenario specific light model} will be built with the help of the \textit{scenario specific heavy model} and the scenario associated samples.
Concretely, we propose a neural architecture search method that can take the budget limitation into consideration, and the knowledge of the \textit{scenario specific heavy model} will be transferred to the light model by employing the distillation philosophy.
In this way, we achieve a \textit{scenario specific light model}, which can satisfy the budget requirement with prominent performance. 
This model will be used for online serving of this scenario.
Subsequently, we will elaborate on the details.

\subsection{Scenario Agnostic Heavy Model}
Thanks to the potential knowledge sharing among these scenarios, we propose to facilitate this system with a \textit{scenario agnostic heavy model}.
This model can be adequately \textit{heavy} if essential as long as better performance can be obtained.
Concretely, to fully ensure the capability of the model, we work on two fronts.

In terms of data utilization, to initialize this model, all available samples will be used. Recall that we may face with the data from $N_0$ scenarios at the beginning, we will collect these samples all together as a new dataset $\mathcal{D}$ for the construction of this model. 
In subsequent, any time when a specific scenario is dealt with, feedback information will be delivered to the \textit{scenario agnostic heavy model}, and the model will be updated to assimilate the knowledge of this specific scenario, which will be detailed in the next subsection.

\begin{figure}[htbp]
	\centering
	\includegraphics[width=0.35\textwidth]{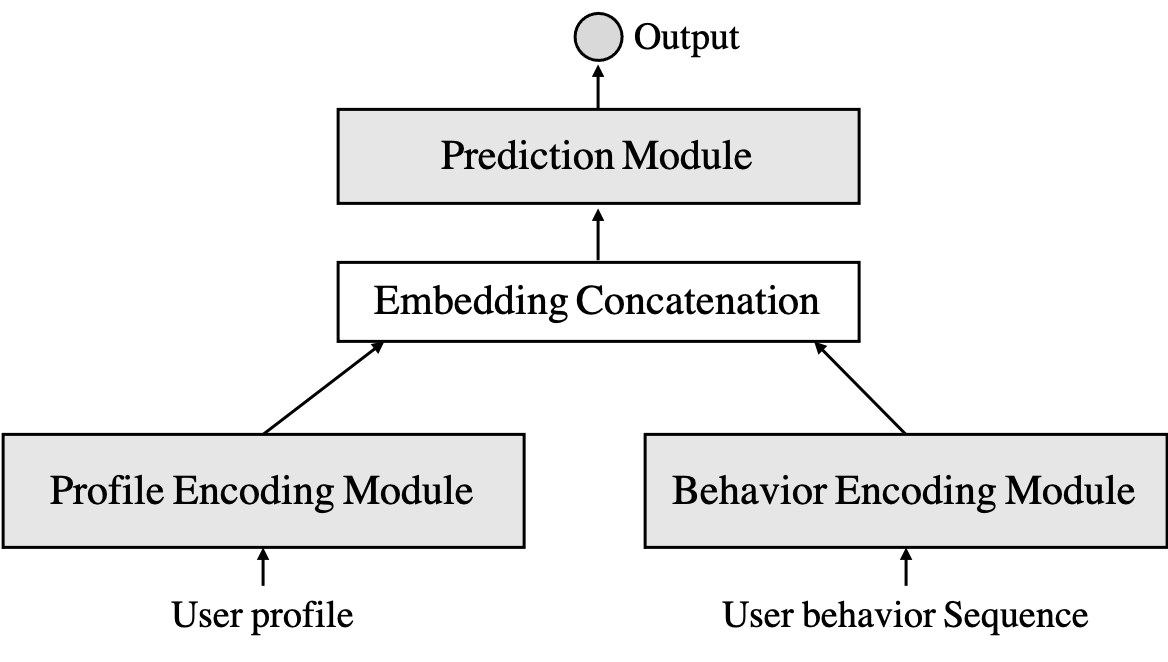}
	\caption{An illustration of the basic architecture of the model.}
	\label{Fig:basic model architecture}
\end{figure}

\begin{figure}[htbp]
	\centering
	\includegraphics[width=0.3\textwidth]{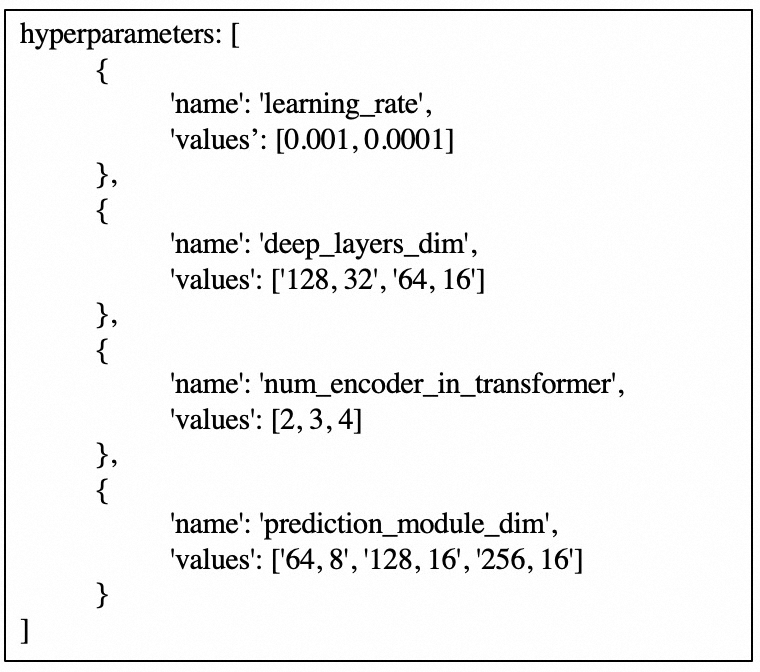}
	\caption{An illustration of the configuration for the pre-designed model search.}
	\label{Fig:hyperparameter demo}
\end{figure}

In terms of model design, we make use of both the exploitation of experts' experience and the exploration of neural architecture search.
For the former, a rough structure is first designed according to experts' experience, and we arm it with a hyperparameter optimization pipeline to obtain the model with more suitable structure and hyperparameters.
As a simplified example, Fig.~\ref{Fig:basic model architecture} shows a basic architecture designed by the expert.
The rough structure is as follows: a profile encoding module 
is used to extract the embedding of the user profile information, a behavior encoding module 
is used to extract the embedding of the user behavior sequence, a prediction module is used after concatenating the two embeddings for final prediction.
For the search of the detailed structure and other hyperparameters, we formalize it as a hyperparameter optimization problem. 
An example of the search space is shown in Fig.~\ref{Fig:hyperparameter demo}, in which the searched hyperparameters include the learning rate, the dimensions of the MLP layers in the profile encoding module, 
the number of encoders used in transformer in the behavior encoding module, and the dimensions of the MLP layers in the prediction module.

Another line of strategy for finding a suitable model is by performing automated model architecture search~\cite{DBLP:conf/iclr/LiuSY19,DBLP:conf/nips/ChangZGMXP19,DBLP:conf/icml/PhamGZLD18,DBLP:conf/nips/Luo0WQCL20,DBLP:conf/pakdd/RenLYZ22}, which has been widely validated as a prominent method for model construction.
In our system, we also take full advantage of this paradigm, and we employ a previously proposed method~\cite{DBLP:conf/pakdd/RenLYZ22} as the default method for automatic model architecture search.
In this subsection, we will omit the details of this part and get into more details when presenting the scenario specific light model.

\begin{figure}[htbp]
	\centering
	\includegraphics[width=0.45\textwidth]{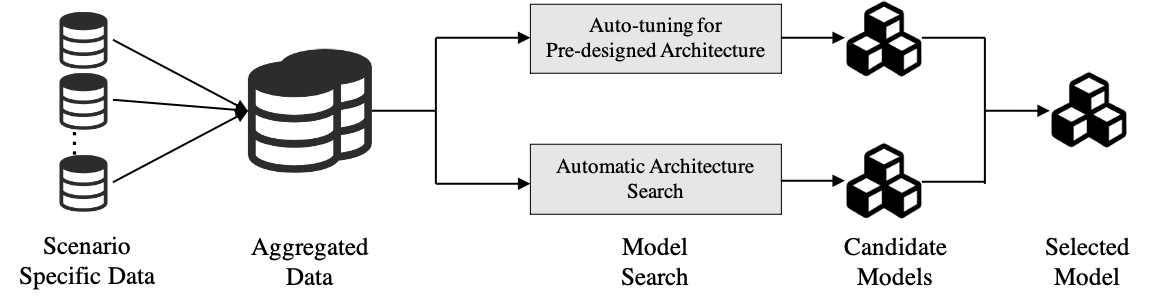}
	\caption{The basic pipeline to initialize the scenario agnostic heavy model.}
	\label{Fig:general model construction}
\end{figure}

In summary, the basic pipeline for building the \textit{scenario agnostic heavy model} is shown in Fig.~\ref{Fig:general model construction}.
After aggregating the data from all available scenarios, two candidate pipelines can be performed. 
The first choice is to get the most suitable specific architecture based on the pre-designed rough architecture by the hyperparameter optimization method, and the other is to perform automatic architecture search to obtain the candidate model.
By choosing the better one from these two candidates, our \textit{scenario agnostic heavy model} is initialized, which will be used and updated throughout the whole framework.
Note that these two pipelines are not necessarily required in practice, the engineers can choose one of them to initialize the model.
We denote this model as $f_0$ with parameter $\theta_0$ for further usage in the subsequent sections.

\subsection{Scenario Specific Heavy Model}
Note that the goal of the \textit{scenario agnostic heavy model} is to leverage the information of all encountered scenarios, and the general model obtained in this way may be incongruous for a specific scenario.
To mitigate this case, when it comes to a specific scenario, the \textit{scenario agnostic heavy model} will be first copied and then fine-tuned with the scenario specific samples.
We call the resulting fine-tuned model a \textit{scenario specific heavy model}.
At the same time, it is also essential that the knowledge of each specific scenario should be also responded to the \textit{scenario agnostic heavy model}. 
To achieve this, a query set is separated from the scenario specific samples. After the optimization of the \textit{scenario specific heavy model}, the loss on the query set will be calculated, which be further used for the update of the scenario agnostic model.

\begin{figure}[htbp]
	\centering
	\includegraphics[width=0.35\textwidth]{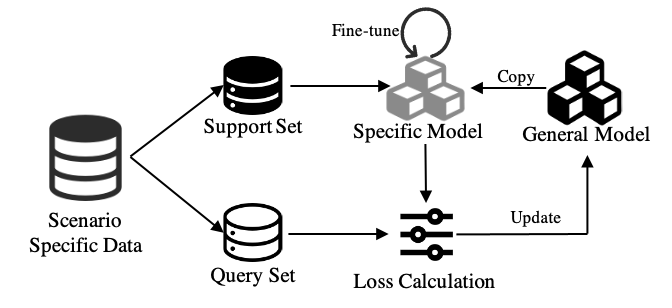}
	\caption{The basic pipeline to obtain the scenario specific heavy model.}
	\label{Fig:specific model construction}
\end{figure}

We provide an illustration in Fig.~\ref{Fig:specific model construction} for a better explanation.
When it comes to a specific scenario $\mathcal{S}_u$, the samples in $\mathcal{D}_u$ are first randomly separated into a support set $\mathcal{D}_u^s$ and a query set $\mathcal{D}_u^q$.
In the meanwhile, a scenario specific model $f_u$ (with parameters $\theta_u$) is first obtained by copying the \textit{scenario agnostic heavy model} $f_0$ (with parameters $\theta_0$), which is further fine-tuned with the samples in the support set $\mathcal{D}_u^s$, i.e.,
\begin{equation}
\small
\theta_u \gets \theta_0 - \gamma \nabla_{\theta_0} \mathcal{L}(\mathcal{D}_u^s, \theta_0) \,,
\label{eq:update_1}
\end{equation}
in which $\mathcal{L}(\mathcal{D}_u^s, \theta_0)$ denotes the loss calculated with parameters $\theta_0$ and data set $\mathcal{D}_u^s$, and $\gamma$ is the learning rate.
After the fine tuning process, the resulting \textit{scenario specific heavy model} $f_u$ ($\theta_u$) will be used for the construction of the \textit{scenario specific light model} described in the subsequent subsection.
And also, the loss is calculated with the \textit{scenario specific heavy model} $f_u$ and the query set $\mathcal{D}_u^q$, which will be further used for the updating of the \textit{scenario agnostic heavy model} $f_0$.
Specifically, the calculation of $\theta_0$ can be illustrated as Eq.~\ref{eq:update_2}.
\begin{equation}
\small
\theta_0 \gets \theta_0 - \eta \nabla_{\theta_0} \mathcal{L}(\mathcal{D}_u^q, \theta_u) \,.
\label{eq:update_2}
\end{equation}

Moreover, note that in practical application, multiple scenarios may be encountered simultaneously, which means that this process will be executed in parallel.
However, one crucial issue is that the update of the \textit{scenario agnostic heavy model} $f_0$ should be properly performed.
In our system, essential supports are equipped for the update of $\theta_0$ with multiple scenarios, this can be simply illustrated as Eq.~\ref{eq:update_3}.
\begin{equation}
\small
\theta_0 \gets \theta_0 - \eta \nabla_{\theta_0} \sum_{u=1}^U \mathcal{L}(\mathcal{D}_u^q, \theta_u) \,,
\label{eq:update_3}
\end{equation}
in which $U$ is the number of simultaneously encountered specific scenarios and $\eta$ is used to control the extent when updating the \textit{scenario agnostic heavy model}. 
We need to mention that in order to prevent the \textit{scenario agnostic heavy model} from being dominated by a subsequent scenario, the parameter $\eta$ is set conservatively in our system.
In practice, more issues need to be considered when building the associated system, which will be further discussed in the subsequent system section.

Furthermore, note that the effectiveness of the \textit{scenario agnostic heavy model} may be critical in real-world tasks, more strategies can be employed to enhance its performance.
For example, we can store all of the training data of all encountered scenarios, and periodically update the parameters of the \textit{scenario agnostic heavy model} with these data. 
This is somehow similar to the Meta-Train procedure in the typical meta learning framework.
These further extensions are also supported in our system.

\subsection{Scenario Specific Light Model}

In real industrial tasks, the time and computing resources are always limited in \textit{inference} stage, while the \textit{scenario specific heavy model} may occupy too many resources. 

Such restrictions require the scenario specific model to achieve comparable performances using fewer computational resources than \textit{scenario specific heavy model}.  
Meanwhile, there are always multiple user behavior sequences in industrial scenarios, which means the behavior encoding module in Fig.~\ref{Fig:basic model architecture} needs to be copied multiple times. 
The behavior encoding module thus contributes most of the computation.
Considering the heaviness of the behavior encoding module, we propose a budget-limited neural architecture search (NAS) method to automatically design \textit{lightweight} sequence model architectures to process user behavior sequences and consequently build a \textit{scenario specific light model}. 
Essentially, in the proposed method, we construct a search space tailored for sequential behavior data and conduct a computational consumption-constrained search algorithm to find the optimal \textit{lightweight} sequence model structure, which will be detailed subquently.
\begin{figure}[htbp]
	\centering
	\includegraphics[width=0.25\textwidth]{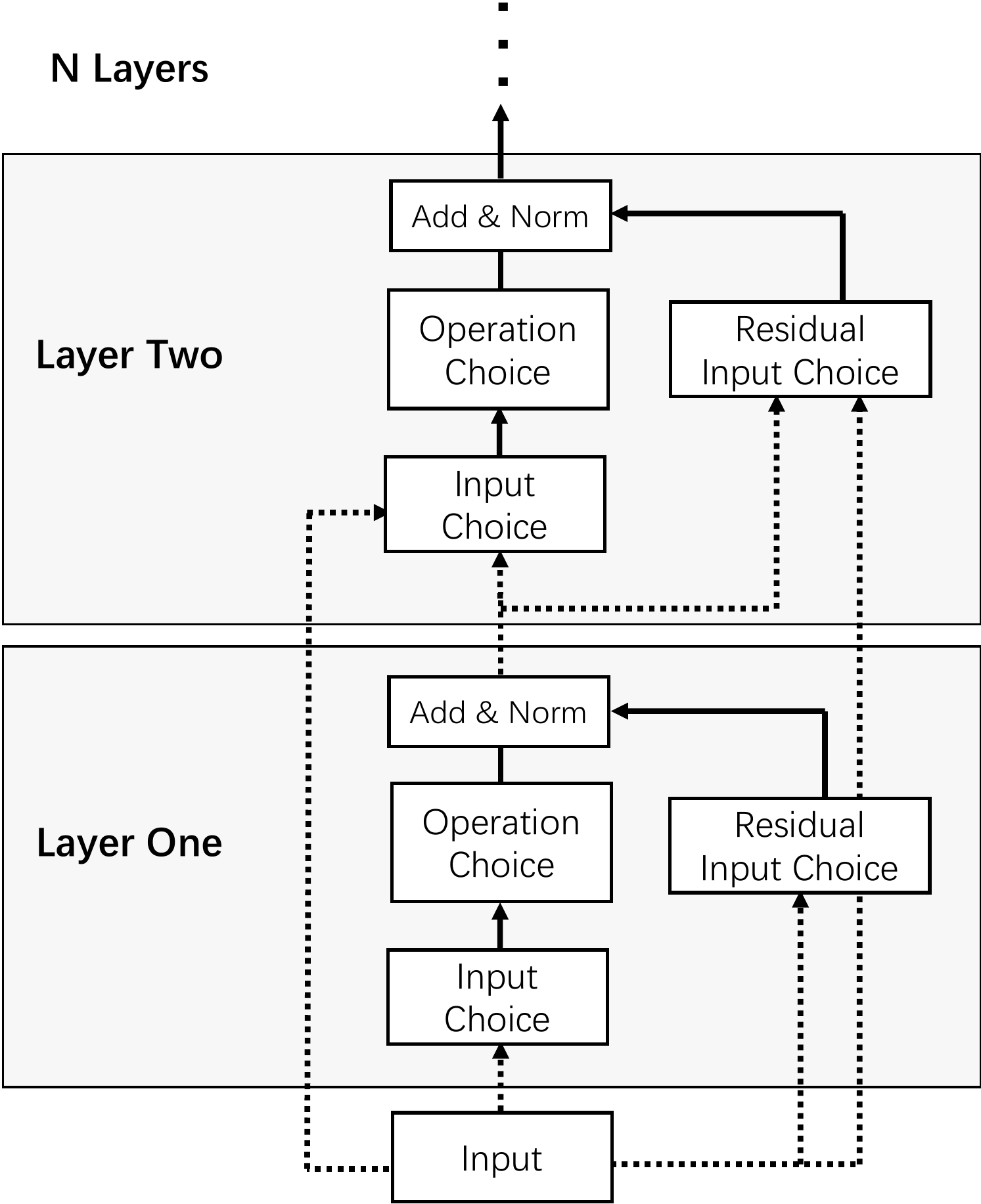}
	\caption{The search space of the behavior encoding module for the  scenario specific light model.}
	\label{Fig:search_space}
\end{figure}

We propose a novel search space tailored for behavior sequence modeling. The overall architecture of the search space is shown in Fig.~\ref{Fig:search_space}. 
We formulate the overall architecture as a stack of $N$ layers. The final output is computed by summing all the layers' outputs attentively. 
For each layer $L$, there are three components that are searchable: 
(i) The input of layer $L$. 
 In each layer, the input can be chosen from all the previous layers' outputs. For example, Layer 3 can
select one from set \{original input, layer 1's output, layer 2's output\} as its input. In this way, the layers can be assembled both in cascade and in parallel.  For example, if layer 2 and layer 1 both choose the original input as their input, they are in parallel. If layer 1 chooses the original input as its input and layer 2 chooses layer 1's output as its input, layer 1 and layer 2 are in cascade.
(ii) The operation to conduct on the input in layer $L$. In each layer, an operation is selected from the candidate operations to extract latent features from the sequence inputs. 
First, we use 1D standard convolutions and dilated
convolutions with kernel size \{1, 3, 5, 7, 9\}. Note
that the 1-D convolution with kernel size = 1 is equivalent to the linear layer. To
keep the shape of output the same as input, we utilize the convolution of stride = 1
with SAME padding and the filter dimensions of output is the same as input.
Second, we include LSTM~\cite{graves2012long} and multi-head self-attention~\cite{vaswani2017attention} in the operation
candidate set to capture global features. 
(iii) The residual inputs of layer $L$. As residual connections are proven useful in deep neural networks~\cite{vaswani2017attention,he2016deep}, we add a residual input choice module to the search space.   A layer can decide whether each one of its former layers’ outputs can be
its residual input independently, which means that one layer
can have multiple residual inputs. For example, the second layer can have two residual inputs both
from the original input and the first layer's output.

After defining the search space, we conduct a computational consumption-constrained search algorithm to find the optimal sequence model structure out of the search space. Following previous works~\cite{xie2018snas,chen2021adabert}, we use the number of floating point operations (FLOPs) as an approximation to the computational resources consumed by the model in the inference phase.
Formally, we denote a neural architecture as $\alpha$ and the
weights of $\alpha$ as $\omega_{\alpha}$. We divide the original train data into train split and validation split.  The goal of NAS is to find an architecture $\alpha$, which can achieve the minimum validation loss $\mathcal{L}_{val}$ on the validation split after the weights $\omega_{\alpha}$ of $\alpha$ being trained by minimizing the train loss $\mathcal{L}_{train}$ on the train split. $\mathcal{L}_{val}$ is shown as Eq.~\ref{eq:lval}.
\begin{align}
    \small
    &\mathcal{L}_{val} =  \mathcal{L}(\alpha, \omega_{\alpha}^{*}, \mathcal{D}_{val})+ \lambda \mathcal{L}_{FLOPs}(\alpha) 
    \nonumber\\
    & \text{s.t. } \omega_{\alpha}^{*}=\text{argmin}_{\omega}\mathcal{L}(\alpha, \omega_{\alpha}, \mathcal{D}_{train}) \,,
    \label{eq:lval}
\end{align}
where $\mathcal{D}_{train}$ is the train split and $\mathcal{D}_{val}$ is the validation split. $\mathcal{L}(\alpha, \omega_{\alpha},\mathcal{D})$ denotes the loss on data split $\mathcal{D}$. $\mathcal{L}_{FLOPs}(\alpha)$ is the normalized number of FLOPs of the neural architecture $\alpha$. 
Such loss guide the optimization process to balance between precision and computational resource cost with the trade-off parameter $\lambda$.

Also, to boost the performance of the searched light model, we manage to distill the knowledge of the scenario specific heavy model. The scenario specific light model is the student model and the scenario specific heavy model serves as the teacher model. We use the hard label $y_{hard}$ of the sample and the output of the scenario specific heavy model as the soft label $y_{soft}$. The loss is computed as Eq.~\ref{eq:dis_loss}.

\begin{equation}
    \small
    \mathcal{L}(\alpha, \omega_{\alpha}, \mathcal{D})= \mathcal{L}_{CE}(y_{hard}^{'},y_{hard}) + \delta \mathcal{L}_{CE}(y_{soft}^{'},y_{soft}) \,,
    \label{eq:dis_loss}
\end{equation}
where $y_{hard}^{'}$ is the student model's hard prediction and $y_{soft}^{'}$ is the student model's soft prediction, and $\delta$ is used to balance these two parts.

Given an $\alpha$, $\omega_{\alpha}^*$ can be easily obtained through gradient descent on the train split as shown in Eq.~\ref{eq:weights}.
\begin{equation}
\small
    \omega_{\alpha} \gets \omega_{\alpha} - \beta\nabla_{\omega_{\alpha}}\mathcal{L}_{train} \,,
    \label{eq:weights}
\end{equation}
where $\beta$ is the learning rate.

\begin{figure*}[htbp]
\centering
 	\includegraphics[width=0.85\textwidth]{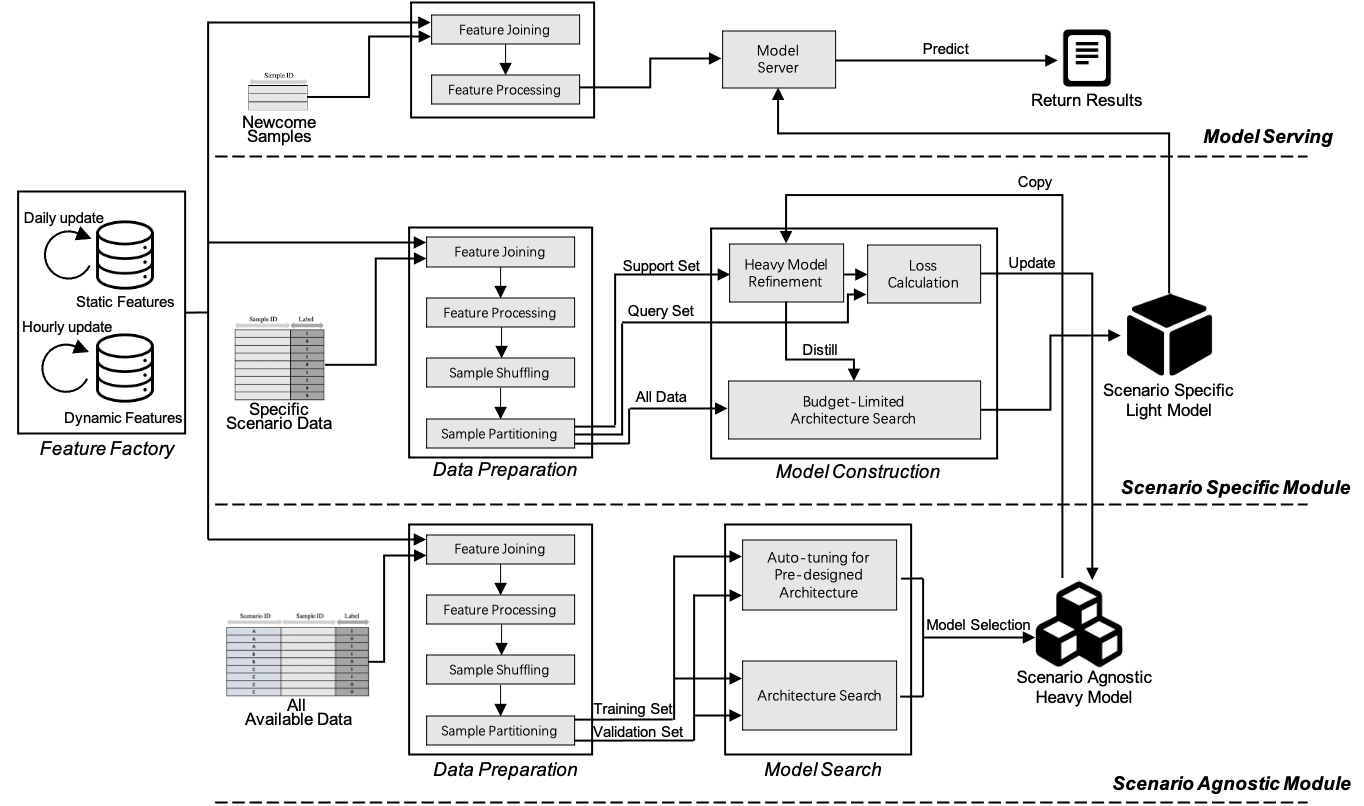}
	\caption{An illustration of the main components of the whole ALT system.}
	\label{Fig:the whole system}
\end{figure*}

However, as $\alpha$ is discrete, $\mathcal{L}_{val}$ is not  differentiable with respect to $\alpha$. 
We follow the operation choice method of GDAS~\cite{dong2019searching}. Suppose the candidate operation set is $\mathcal{O}=\{\mathcal{O}_1,\mathcal{O}_2,\cdots,\mathcal{O}_m,\cdots,\mathcal{O}_{n_o}\}$ and $|\mathcal{O}|=n_o$ , we sample architectures from a distribution characterized by a set of  learnable architecture distribution weights $[o_1, o_2, \cdots, o_{n_o}]$. To enable back-propagation, we replace the sampling process
with the Gumbel Softmax trick:
\begin{equation}
    \small
    P_{\mathcal{O}_i}=\frac{\text{exp}((o_i+g_i)/\tau)}{\sum_{j=1}^{j=n_o}[ \text{exp}((o_j+g_i)/\tau)]} \,,
    \label{eq:prob}
\end{equation}
where $g_i$ is a random noise drawn from $\text{Gumbel}(0, 1)$ distribution, $\tau$ is the temperature, when $\tau \rightarrow 0$, the  distribution $[P_{\mathcal{O}_1}, P_{\mathcal{O}_2},\cdots,P_{\mathcal{O}_{n_0}}]$ approaches one-hot, thus the operation with probability 1 is sampled. Similar to GDAS, the output $y_0$ of the sampled operation $\mathcal{O}_m$ can be computed as follows:
\begin{align}
    \small
    &y_0(x) = (1-\text{detached}(P_{\mathcal{O}_m})+P_{\mathcal{O}_m}) \cdot f_{\mathcal{O}_m}(x) 
    \nonumber\\
    &\text{s.t. } \mathcal{O}_m = \mathop{\arg\max}\limits_{\mathcal{O}_i} P_{\mathcal{O}_i} \,,
\end{align}
where $x$ is the input and $\text{detached}(a)$ means $a$ does not receive gradients.
In this way, the output is $f_{\mathcal{O}_m}(x)$ during forward and we can backpropagate to the learnable weight $o_m$ during back-propagation. After training, each operation will have a probability $P_{\mathcal{O}_i}^{'}$:
\begin{equation}
\small  
    P_{\mathcal{O}_i}^{'}=\frac{\text{exp}(o_i)}{\sum_{j=1}^{j=n_o}[ \text{exp}(o_j)]} \,.
    \label{eq:orig_prob}
\end{equation}
The input and residual input search can be done in the same way as the operation search, which will not be detailedly specified due to the space limitations.
To find the best neural architecture under the given FLOPs constraint, we select the operation combination which satisfies the FLOPs constraint and has the maximum joint probability.

One issue that needs to be mentioned is that additional time cost may be involved when using NAS based method.
However, as we addressed, for our concerned setting and most of the related real-world tasks, the additional time cost in the training stage is acceptable, and our main concern of the time and resource cost is typically about the inference stage.

\section{System}
\label{system}

\subsection{Overview}

In this section, we elaborate on the implementation details from a systematic perspective.
To build a sound system, in addition to the algorithm ability mentioned previously, many other modules should be equipped.
Concretely, as demonstrated in Fig.~\ref{Fig:the whole system}, several essential modules are included in our system, and we will roughly introduce the following components.
\begin{itemize}
\item Feature Factory: To store and maintain the features that are used in the subsequent models. The Feature Factory module, together with the Data Preparation module, forms a mature data preparation and processing pipeline.
\item Scenario Agnostic Module: To initialize and maintain the \textit{scenario agnostic heavy model}. With the help of auto-tuning for pre-designed architecture and NAS techniques, an effective model is generated and maintained to leverage the information from all associated scenarios.
\item Scenario Specific Module: To refine the \textit{scenario specific heavy model} and construct the \textit{scenario specific light model}. By employing the budget-limited NAS method and distilling the knowledge of the refined \textit{scenario specific heavy model}, an effective and light model is constructed for online serving.
\item Model Serving: To provide efficient and effective online serving. Based on the feature process pipeline and the light model, online serving can be efficiently obtained.
\end{itemize}

Subsequently, we will detail these crucial modules, with the system level support briefly depicted.

\subsection{Feature Factory}
We take full advantage of the data processing and storage ability of \textbf{MaxCompute}\footnote{https://www.alibabacloud.com/zh/product/maxcompute} to design the Feature Factory module.
As shown in Fig.~\ref{Fig:basic model architecture}, the features that we used in the models can be roughly separated into two groups, i.e., the user profile part, which is relatively stable, and the user behavior part, which is frequently updated.
To accommodate this, different update frequencies are employed in the Feature Factory, so that, on one hand, the effectiveness of the related features (e.g., the behavior sequence) can be guaranteed, on the other hand, the computing resources can be invoked efficiently.
Representatively, the stable user profile features can be updated daily or even monthly, while the dynamic behavior sequence needs to be updated hourly or more often in our system. 
The feature update process will be regularly scheduled to support the modeling process. 

When a new model construction procedure proceeds, the Data Preparation module can be executed based on the prepared feature factory, which typically includes the following procedure: feature joining that links the users with the related features, feature processing that performs the necessary feature processing like normalization and discretization, sample shuffling that shuffles the sample order if needed, sample partitioning that separates the samples into different subsets if needed.
After these processes, the model construction procedure and online predicting procedure can be subsequently performed with the prepared samples.

\subsection{Scenario Agnostic Module}
As a crucial part, the construction and maintenance of the \textit{scenario agnostic heavy model} are emphasized in our system.
Note that auto-tuning of the pre-designed architecture is used in our system, for which ingenious system implementation for the hyperparameter optimization module is realized.

\begin{figure}[hbtp]
\centering
 	\includegraphics[width=0.3\textwidth]{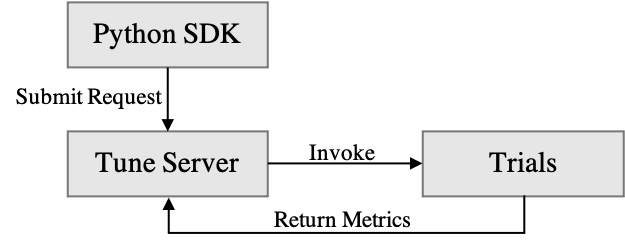}
	\caption{The high-level demonstration of AntTune.}
	\label{Fig:AntTune}
\end{figure}

Specifically, a hyperparameter optimization module named \textbf{AntTune}~\cite{zhou2023anttune}
is employed, and the high-level demonstration of it is shown in Fig.~\ref{Fig:AntTune}.
Roughly speaking, when a new hyperparameter optimization task is faced, the SDK will first submit the request to the tune server with the necessary information such as the search space and the time limitation, etc.
The server will effectively generate candidate trials which will be distributedly executed, i.e., building the model with the specific hyperparameter configuration, and then the metrics will be returned to the server, so that the new population can be generated accordingly and the best model can be selected finally.
Note that various hyperparameter optimization methods, such as evolutionary algorithm~\cite{DBLP:journals/ec/HansenMK03}, Bayesian optimization~\cite{DBLP:conf/nips/SnoekLA12}, and classification-based strategy~\cite{DBLP:conf/aaai/YuQH16}, etc., have been implemented in our system, and we use RACOS~\cite{DBLP:conf/aaai/YuQH16} as the default algorithm due to its high efficiency and appealing flexibility~\cite{DBLP:conf/ijcai/QianHY16,DBLP:conf/aaai/HuQY17,DBLP:conf/intellisys/ZhangSLYLZ21,DBLP:conf/ecai/HuLYYL20}.
It is worth noting that in our implementation of the hyperparameter optimization module, several systematic optimization tricks are involved, which include the time limitation for every single trial and the overall job, early-stopping for the futureless trials, and the fault-tolerant mechanism, etc. 
We will omit the tedious details of these parts.

This hyperparameter optimization module ensures efficient search for the pre-designed architecture, whose resulting candidate model will be compared with the automatically constructed candidate to decide the \textit{scenario agnostic heavy model}, as elaborated in the method section.

\subsection{Scenario Specific Module}
As illustrated in Fig.~\ref{Fig:the whole system}, in this module, the \textit{scenario specific heavy model} is first generated by copying the \textit{scenario agnostic heavy model} as initialization, it will then be refined with the scenario specific samples.
It is no doubt that the \textit{scenario agnostic heavy model} is a promising initialization, and an effective model can be then obtained efficiently based on it.
After the optimization of the \textit{scenario specific heavy model}, two matters are then performed, i.e., sending the feedback information back to update the \textit{scenario agnostic heavy model}, and building the \textit{scenario specific light model} by using the scenario specific samples and distilling the knowledge from the \textit{scenario specific heavy model}.

For the system implementation, the situation where multiple scenarios may come simultaneously is addressed. 
To adapt to this situation, the ability to handle various scenarios in parallel is equipped.
One issue that is worth noting is the proper update for the \textit{scenario agnostic heavy model} when simultaneously handling multiple scenarios.
In our system, this situation is fully considered, and the asynchronous update is supported to accommodate the difference in computational efficiency.

\subsection{Model Serving}
The obtained \textit{scenario specific light model} will be automatically uploaded and deployed to \textbf{model server} for online serving.
In the online serving stage, thanks to the prepared Feature Factory module and the Data Preparation module, the features of the coming users can be efficiently joined and processed when they come to the specific scenario.
After that, the prediction results can be efficiently obtained with the associated light model and delivered for further decision in the subsequent stages.
\section{Experiment}
\label{experiment}
In this section, a series of experiments are conducted based on real-world datasets to validate the effectiveness of the presented system.

\begin{table*}[htb]
\centering
\caption{The detailed sample size of each scenario for dataset A.}
\label{Tab:data-details}
\scalebox{0.81}{
\begin{tabular}{c|c|c|c|c|c|c|c|c|c|c|c|c|c|c|c|c|c|c}
\toprule
 ID       & 1 & 2 & 3 & 4 & 5 & 6 & 7 & 8 & 9 & 10 & 11 & 12 & 13 & 14 & 15 & 16 & 17 & 18 \\ \midrule
Size    & 1202739  & 930438  & 890908 & 875692  &  530441 & 242858  & 93892  & 88084  & 84466  & 69647  & 62134  & 61869  & 61214  & 51506  & 47219  & 46596  & 28643  & 19973  \\ 
\bottomrule
\end{tabular}
}
\end{table*}

\begin{table*}[htb]
\centering
\caption{The detailed sample size of each scenario for dataset B.}
\label{Tab:data-details-2}
\scalebox{0.81}{
\begin{tabular}{c|c|c|c|c|c|c|c|c|c|c|c|c|c|c|c|c}
\toprule 
ID      & 1 & 2 & 3 & 4 & 5 & 6 & 7 & 8 & 9 & 10 & 11 & 12 & 13 & 14 & 15  \\ \midrule
Size     & 221003 & 139043 & 122863 & 113160 & 103506 & 102792 & 97333 & 91394 & 79890 & 60877 & 60731 & 54548 & 45570 & 43615 & 32893  \\ \midrule
ID       & 16  & 17 & 18 & 19 & 20 & 21 & 22 & 23 & 24 & 25 & 26 & 27 & 28 & 29 & 30  \\ \midrule
Size    & 30505 & 26861 & 22340 & 17256 & 16294 & 13108 & 12143 & 7677 & 4825 & 4321 & 3430 & 2870 & 1574 & 976 & 493   \\ 

\bottomrule
\end{tabular}
}
\end{table*}

\subsection{Experimental Setup}

\subsubsection{Dataset Description}
We perform our experiments based on the data collected from two real-world applications.
\textbf{Dataset A} is collected based on the a risk control task\footnote{Some details are omitted due to commercial confidentiality here and below.}, in which several banks are involved, and each may be with different distributions of users.
The goal is to predict the probability of default for each loan application.
Concretely, there are $18$ participants in the task, and the details of the sample size information are concluded in Table~\ref{Tab:data-details}.
The features in this task mainly include two parts, i.e., the basic profiles with $69$ attributes and the user behavior sequences with a maximal length of $128$.
\textbf{Dataset B} is collected based on an advertising task, in which $32$ scenarios are involved. The goal herein is to pick out proper potential users for the advertisers.
Similarly, the features consist of a basic profile vector with $104$ attributes and a user behavior sequence with a maximal length of $128$.
The sample size information is shown in Table~\ref{Tab:data-details-2}.

In our experiments, we randomly separate $20\%$ of the samples as the test set and keep the rest as the training set for model training. 
Moreover, to simulate the setting in this paper, for both datasets only $8$ scenarios are randomly selected as the initial scenarios, and the rest will come subsequently. This means that only the samples of the selected $8$ scenarios will be used for initializing the scenario agnostic heavy model.

\subsubsection{Compared Methods}
Several strategies for building the models are compared to provide a demonstration of the effectiveness of the proposed framework.
To simplify the comparison, we first set some pre-defined (heavy/light) models and compare the performance of these models under different training strategies. 
The details of the compared strategies are concluded as follows.
\begin{itemize}
    \item \textbf{Single-Heavy} (\textbf{SinH}). The pre-defined heavy model is trained with the data from each specific scenario, and this model is further evaluated.
    \item \textbf{Meta-Heavy} (\textbf{MeH}). The pre-defined heavy model is first trained with the $8$ initial scenarios. Then for each specific scenario (including both initial scenarios and the subsequent scenarios), the model is fine-tuned with the scenario specific data, and the feedback information is sent back to update the pre-defined heavy model. This is somehow similar to the process in traditional meta-learning based frameworks. The fine-tuned heavy model is evaluated in this strategy. 
    \item \textbf{Meta-Light} (\textbf{MeL}). The pre-defined heavy model is first trained with the $8$ initial scenarios and fine-tuned as in the MeH strategy. Moreover, the pre-defined light model is trained with the scenario specific data, with the knowledge distilled from the fine-tuned heavy model.
    \item \textbf{Ours}. The pre-defined heavy model is first trained with the 8 initial scenarios and fine-tuned as in the MeH strategy. Moreover, a light model is searched with the proposed budget-limited NAS method. The only difference of this strategy from the MeL strategy is that the light model is obtained by the proposed NAS method rather than the pre-defined architecture. Note that the selected architecture is the one with the maximal joint probability among the candidates that satisfies the FLOPs constraint. 
    Here, the upper bound of the FLOPs for the searched architectures is set to be the same as the light models.
\end{itemize}

\subsubsection{Implementation Details}
All models that were experimented with in this work are designed following the basic architecture as shown in Fig.~\ref{Fig:basic model architecture}.
For the profile encoding module, a deep layer module is simply used and fixed for all aforementioned models.
As for the behavior encoding module, we separately tests with LSTM based architecture~\cite{DBLP:journals/neco/HochreiterS97} and BERT based architecture~\cite{DBLP:conf/naacl/DevlinCLT19}. 
Note that a heavy architecture and a light architecture are involved in our proposed framework. 
Concretely, for the heavy architecture, we use 6-layer LSTMs and 6-layer BERTs.
For the light architecture, we use 3-layer LSTMs and 3-layer BERTs.
The hidden unit number of one LSTM layer is 15. Each BERT layer has 15 hidden units and 32 intermediate hidden units.
Moreover, for the budget NAS part, the candidates are as follows:
1D standard convolutions and dilated
convolutions with kernel sizes \{1, 3, 5, 7\}, 1D average pooling and 1D max pooling with kernel size 3, LSTM and self-attention module.

All models are optimized with Adam~\cite{DBLP:journals/corr/KingmaB14}, with a learning rate of $0.001$ and a batch size of $512$. 
We run for $5$ epochs for each dataset, and the cross-entropy loss is minimized as the main loss term.

\subsubsection{Evaluation Metrics}
The evaluation metrics in this work are two folds. 
On the one hand, the effectiveness of the trained models is evaluated with AUC (Area under the ROC curve) score.
On the other hand, the efficiency of the models is evaluated with the FLOPs (floating point operations per second) information and the inference time.

\subsection{Experimental Results and Analyses}

\begin{table}[htbp]
\centering
\caption{The AUC scores of the compared methods for dataset A.}
\label{Tab:result-AUC}
\scalebox{0.9}{ 
\begin{tabular}{c|cccc|cccc}
\toprule
\multirow{2}{*}{} & \multicolumn{4}{c|}{LSTM-based} & \multicolumn{4}{c}{BERT-based} \\ \cmidrule{2-9} 
                  & SinH   & MeH   & MeL   & Ours   & SinH   & MeH   & MeL   & Ours   \\ \midrule
1	&	0.657 	&	0.683 	&	0.657 	&	0.678 	&	0.663 	&	\textbf{0.687} 	&	0.665 	&	0.685 	\\
2	&	0.744 	&	0.745 	&	0.744 	&	0.737 	&	0.743 	&	0.744 	&	0.736 	&	\textbf{0.746} 	\\
3	&	0.751 	&	0.754 	&	0.754 	&	0.760 	&	0.755 	&	0.761 	&	0.755 	&	\textbf{0.766} 	\\
4	&	0.817 	&	0.824 	&	0.813 	&	0.821 	&	0.822 	&	0.826 	&	0.818 	&	\textbf{0.830} 	\\
5	&	0.725 	&	0.725 	&	0.725 	&	0.729 	&	0.727 	&	0.730 	&	0.726 	&	\textbf{0.733} 	\\
6	&	0.785 	&	0.798 	&	0.775 	&	0.798 	&	0.799 	&	0.798 	&	0.798 	&	\textbf{0.799} 	\\
7	&	0.886 	&	0.896 	&	0.884 	&	0.891 	&	0.891 	&	0.895 	&	0.890 	&	\textbf{0.898} 	\\
8	&	0.724 	&	\textbf{0.726} 	&	0.715 	&	0.715 	&	0.717 	&	0.724 	&	0.713 	&	0.718 	\\
9	&	0.762 	&	0.759 	&	0.761 	&	0.762 	&	0.758 	&	0.762 	&	0.760 	&	\textbf{0.763} 	\\
10	&	0.658 	&	0.664 	&	0.652 	&	0.667 	&	0.659 	&	0.667 	&	0.663 	&	\textbf{0.668} 	\\
11	&	0.661 	&	0.673 	&	0.654 	&	0.672 	&	0.654 	&	\textbf{0.678} 	&	0.660 	&	0.675 	\\
12	&	0.818 	&	0.824 	&	0.822 	&	0.821 	&	0.821 	&	\textbf{0.831} 	&	0.825 	&	0.830 	\\
13	&	0.671 	&	0.676 	&	0.663 	&	0.675 	&	0.672 	&	0.678 	&	0.679 	&	\textbf{0.682} 	\\
14	&	0.686 	&	0.688 	&	0.684 	&	0.712 	&	0.694 	&	\textbf{0.722} 	&	0.687 	&	0.699 	\\
15	&	0.797 	&	0.802 	&	0.799 	&	0.803 	&	0.801 	&	\textbf{0.803} 	&	0.801 	&	0.802 	\\
16	&	0.702 	&	0.704 	&	0.703 	&	0.700 	&	0.694 	&	0.704 	&	\textbf{0.709} 	&	0.706 	\\
17	&	0.661 	&	0.695 	&	0.672 	&	0.683 	&	0.661 	&	\textbf{0.704} 	&	0.684 	&	0.697 	\\
18	&	0.864 	&	0.874 	&	0.853 	&	0.870 	&	0.872 	&	\textbf{0.887} 	&	0.850 	&	0.872 	\\ \midrule
AVG	&	0.743 	&	0.751 	&	0.741 	&	0.750 	&	0.745 	&	\textbf{0.756} 	&	0.746 	&	0.754 	\\ \bottomrule
\end{tabular}
}
\end{table}

\begin{table}[htbp]
\centering
\caption{The AUC scores of the compared methods for dataset B.}
\label{Tab:result-AUC-2}
\scalebox{0.9}{ 
\begin{tabular}{c|cccc|cccc}
\toprule
\multirow{2}{*}{} & \multicolumn{4}{c|}{LSTM-based} & \multicolumn{4}{c}{BERT-based} \\ \cmidrule{2-9} 
                  & SinH   & MeH   & MeL   & Ours   & SinH   & MeH   & MeL   & Ours   \\ \midrule
1	&	0.901 	&	\textbf{0.912} 	&	0.896 	&	0.905 	&	0.903 	&	0.910 	&	0.899	&	0.907	 \\
2	&	0.913 	&	0.925 	&	0.910 	&	0.928 	&	0.911 	&	0.929 	&	0.911	&	\textbf{0.931}	 \\
3	&	0.907 	&	0.919 	&	0.901 	&	0.919 	&	0.906 	&	\textbf{0.924} 	&	0.901	&	0.921	 \\
4	&	0.891 	&	0.902 	&	0.884 	&	\textbf{0.905} 	&	0.898 	&	0.901 	&	0.889	&	0.902	 \\
5	&	0.899 	&	0.911 	&	0.887 	&	0.913 	&	0.893 	&	0.913 	&	0.889	&	\textbf{0.914}	 \\
6	&	0.894 	&	0.902 	&	0.891 	&	0.899 	&	0.896 	&	\textbf{0.909} 	&	0.893	&	0.903	 \\
7	&	0.928 	&	0.944 	&	0.920 	&	0.938 	&	0.927 	&	\textbf{0.945} 	&	0.921	&	0.940	 \\
8	&	0.874 	&	\textbf{0.894} 	&	0.870 	&	0.886 	&	0.881 	&	0.888 	&	0.869	&	0.883	 \\
9	&	0.806 	&	0.824 	&	0.808 	&	0.813 	&	0.804 	&	0.823 	&	0.810 	&	\textbf{0.825} 	 \\
10	&	0.906 	&	0.916 	&	0.909 	&	0.911 	&	0.909 	&	0.920 	&	0.914 	&	\textbf{0.922} 	 \\
11	&	0.863 	&	0.883 	&	0.868 	&	0.881 	&	0.869 	&	\textbf{0.891} 	&	0.871 	&	0.891 	 \\
12	&	0.862 	&	0.880 	&	0.861 	&	0.880 	&	0.860 	&	\textbf{0.893} 	&	0.861 	&	0.891 	 \\
13	&	0.850 	&	0.868 	&	0.860 	&	0.868 	&	0.858 	&	0.869 	&	0.864 	&	\textbf{0.872} 	 \\
14	&	0.856 	&	\textbf{0.873} 	&	0.863 	&	0.869 	&	0.863 	&	0.871 	&	0.866 	&	0.872 	 \\
15	&	0.725 	&	0.743 	&	0.733 	&	0.739 	&	0.729 	&	\textbf{0.743} 	&	0.737 	&	0.742 	 \\
16	&	0.704 	&	0.724 	&	0.708 	&	0.719 	&	0.703 	&	0.727 	&	0.704 	&	\textbf{0.729} 	 \\
17	&	0.725 	&	\textbf{0.740} 	&	0.729 	&	0.737 	&	0.729 	&	0.739 	&	0.733 	&	0.739 	 \\
18	&	0.625 	&	0.648 	&	0.624 	&	0.639 	&	0.628 	&	\textbf{0.652} 	&	0.628 	&	0.649 	 \\
19	&	0.775 	&	\textbf{0.800} 	&	0.789 	&	0.799 	&	0.773 	&	0.798 	&	0.784 	&	0.789 	 \\
20	&	0.688 	&	0.716 	&	0.695 	&	0.707 	&	0.696 	&	\textbf{0.718} 	&	0.699 	&	0.714 	 \\
21	&	0.721 	&	0.751 	&	0.714 	&	0.734 	&	0.710 	&	\textbf{0.760} 	&	0.726 	&	0.745 	 \\
22	&	0.757 	&	0.777 	&	0.758 	&	0.770 	&	0.762 	&	\textbf{0.778} 	&	0.760 	&	0.770 	 \\
23	&	0.769 	&	0.784 	&	0.766 	&	0.775 	&	0.768 	&	\textbf{0.789} 	&	0.766 	&	0.785 	 \\
24	&	0.718 	&	\textbf{0.750} 	&	0.725 	&	0.739 	&	0.720 	&	0.748 	&	0.720 	&	0.733 	 \\
25	&	0.747 	&	\textbf{0.779} 	&	0.755 	&	0.764 	&	0.761 	&	0.770 	&	0.749 	&	0.760 	 \\
26	&	0.664 	&	0.692 	&	0.670 	&	0.677 	&	0.672 	&	\textbf{0.699} 	&	0.671 	&	0.683 	 \\
27	&	0.749 	&	\textbf{0.777} 	&	0.750 	&	0.768 	&	0.750 	&	0.770 	&	0.755 	&	0.772 	 \\
28	&	0.633 	&	0.670 	&	0.641 	&	0.659 	&	0.640 	&	\textbf{0.680} 	&	0.649 	&	0.659 	 \\
29	&	0.643 	&	0.676 	&	0.649 	&	0.656 	&	0.652 	&	\textbf{0.682} 	&	0.659 	&	0.676 	 \\
30	&	0.646 	&	0.691 	&	0.660 	&	0.682 	&	0.652 	&	\textbf{0.696} 	&	0.665 	&	0.678 	 \\
31	&	0.712 	&	0.743 	&	0.729 	&	0.739 	&	0.719 	&	\textbf{0.749} 	&	0.731 	&	0.736 	 \\
32	&	0.721 	&	0.759 	&	0.733 	&	0.750 	&	0.726 	&	\textbf{0.763} 	&	0.731 	&	0.759 	 \\ \midrule
avg	&	0.784 	&	0.805 	&	0.786 	&	0.799 	&	0.786 	&	\textbf{0.808} 	&	0.788 	&	0.803 	 \\
\bottomrule
\end{tabular}
}
\end{table}

\begin{table}[htbp]
\centering
\caption{The averaged FLOPs information and inference time of different models.}
\label{Tab:result-FLOPs}
\scalebox{0.78}{ 
\begin{tabular}{c|c|ccc|ccc}
\toprule
                                                                                   &           & \multicolumn{3}{c|}{LSTM-Based} & \multicolumn{3}{c}{BERT-Based} \\ \cmidrule{3-8} 
                                                                                   &           & Heavy     & Light    & Ours     & Heavy    & Light    & Ours     \\ \midrule
\multirow{2}{*}{\begin{tabular}[c]{@{}c@{}}FLOPs\\ Information\end{tabular}} & Dataset A & 4.78M     & 2.46M    & 2.12M    & 4.74M    & 2.44M    & 2.07M    \\
                                                                                   & Dataset B &  5.19M &  2.75M        &  2.61M  &   5.14M  & 2.68M     &   2.58M  \\ \midrule
\multirow{2}{*}{\begin{tabular}[c]{@{}c@{}}Inference \\ Time\end{tabular}}  & Dataset A &  10.25ms    &  5.14ms  &   3.13ms  &  6.71ms   &    3.42ms   &  2.96ms  \\
                                                                                   & Dataset B &  11.12ms &  5.43ms    &    2.61ms      &  7.29ms   &   3.72ms &  3.54ms     \\ \bottomrule
\end{tabular}
}
\end{table}

\subsubsection{Main Results}
The main results regarding the AUC metric are concluded in Table~\ref{Tab:result-AUC} and Table~\ref{Tab:result-AUC-2}
with the best marked in bold, and the averaged FLOPs information and the inference time information of the related models are presented in Table~\ref{Tab:result-FLOPs}.
As we can see, for both datasets, the best scores are always achieved by the MeH strategy or our proposed method.
That's reasonable, since MeH is with sufficient model complexity, and makes full use of data from relevant scenarios.
And thanks to the distillation strategy and the proposed NAS method, our strategy can result in a much lighter model than the heavy model, while the performance can be competitive with the MeH strategy, and be much better than the pre-defined light model.
One interesting phenomenon is that in some scenarios SinH can perform better than MeL, while in some others MeL behaves better than the SinH strategy.
For the former, one reason might be that improving the complexity of the model is suitable for these scenarios, while for the latter, one possible explanation is that the gain from distillation is greater than the gain from increasing model complexity.

\subsubsection{The Benefits of Leveraging All Data Information}
When focusing on the performance of MeH and SinH in Table~\ref{Tab:result-AUC} and Table~\ref{Tab:result-AUC-2}, we can evidently find out the benefits of leveraging the related data for the modeling of each specific scenario.
Almost all of the scenarios are benefited from the use of other related scenarios.
For those scenarios where the performances are not evidently improved, one possible reason is that this scenario is a little distant from the other scenarios, which means that the information that can be transferred among them is somehow restricted.
Still, the performance of these scenarios will not obviously deteriorate under our framework.

\subsubsection{The Benefits of the Budget-limited NAS Method}

The goal of the budget-limited NAS method is to obtain an effective yet light model.
Through our experiments, we can see that the obtained scenario specific model is much lighter than the heavy model, and the corresponding inference time is much shorter than the heavy model(see the details in Table~\ref{Tab:result-FLOPs}), while the performance is pretty competitive, and can perform even better in some scenarios (see the details in Table~\ref{Tab:result-AUC} and Table~\ref{Tab:result-AUC-2}).
Moreover, when we compare the performance of the light model obtained from the NAS method with the pre-defined light model, we can obviously observe the superiority of our method from the perspective of both effectiveness and efficiency.

\subsubsection{Illustration of the Searched Architectures}

\begin{figure}[ht]
    \centering
    \subfigure[]{
        \label{Fig:illustration-search-architecture-a}
        \includegraphics[width=0.22\textwidth]{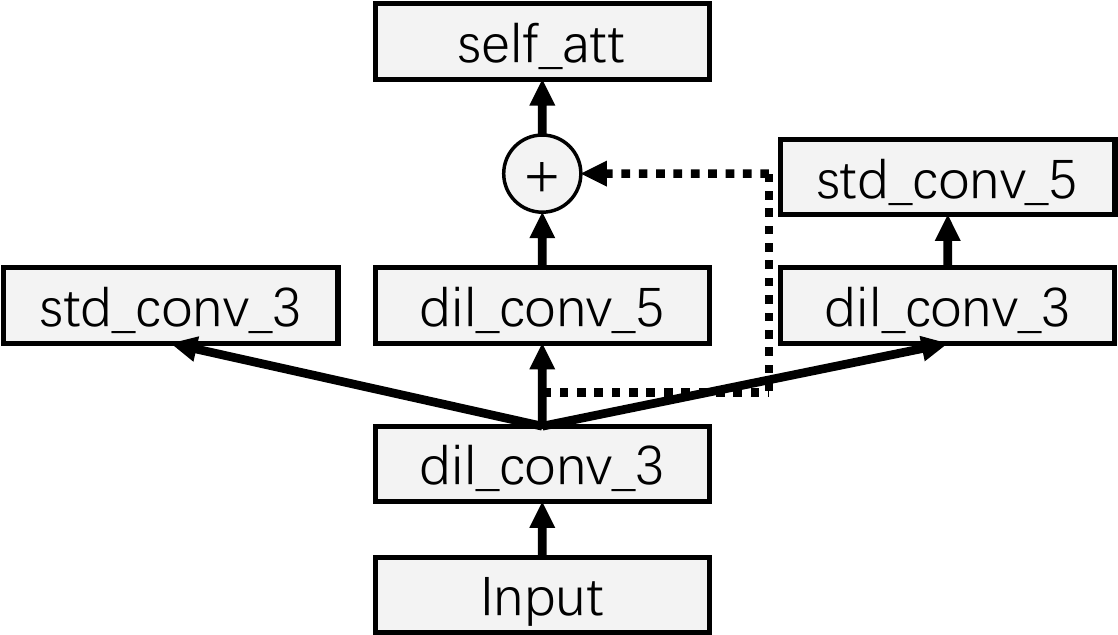}
    }
    \subfigure[]{
        \label{Fig:illustration-search-architecture-b}
        \includegraphics[width=0.22\textwidth]{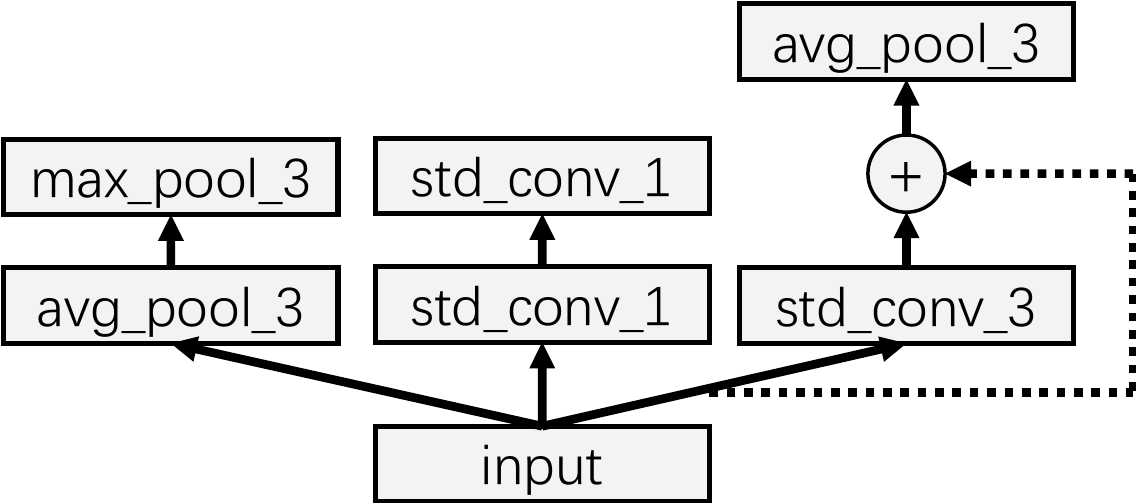}
    }
    \caption{The illustration of the searched architectures. Solid lines represent dataflow to layer input, and dotted lines indicate residual connection. The final output is computed by summing all the layers' outputs attentively. }
    \label{Fig:illustration-search-architecture}
\end{figure}

To better understand the NAS method, we also provide two representatives of the searched architectures from dataset A in Fig.~\ref{Fig:illustration-search-architecture}. 
In Fig.~\ref{Fig:illustration-search-architecture-a}, the searched architecture $a$ for Scenario $4$ with a relatively large sample size is presented, while the searched architecture $b$ for Scenario $15$ with a relatively small sample size is shown in Fig.~\ref{Fig:illustration-search-architecture-b}. 
We can observe that architecture $a$ is more complicated than architecture $b$, as the average filter size is bigger in architecture $a$ and there are more trainable parameters in architecture $a$. 
This may be because there's much more data in Scenario $a$, and the searched architecture $a$ needs to be more complicated to fit more training data.
Also, these illustrations show us the superiority of the proposed NAS method compared to the pre-defined architectures, since the NAS method can search for a more suitable architecture to flexibly adapt to specific data characteristics.

\subsubsection{The Benefits of NAS for Model Initialization}
Note that in our system, further efforts can be taken to obtain a better initial general model, by employing the NAS approach or hyperparameter optimization techniques.
We also conduct some additional experiments with dataset A to verify the benefits of using further techniques when initializing the general model.
Specifically, we perform model construction and validate the performance with different numbers of initial scenarios. 
We take the models obtained by the NAS method and the pre-defined LSTM/BERT-based heavy architecture into comparison.
Note that the initial scenarios are randomly selected, and the evaluation is performed on a leave-out validation set sampled from all initial scenarios.
We repeat this process three times, and report the averaged results herein.
Table~\ref{Tab:result-initial-model} concludes the averaged results of the compared methods with different numbers of initial scenarios.
As we can see, with the help of the NAS method, the performance of the initial general model can be further improved.
For some situations, the improvements are relatively significant, while for some others, the improvements are relatively small. 
Still, these results show us the potential helpfulness of the strategies in the scenario agnostic module since the general model is the base of all of the subsequent scenario specific models.
The improvement of this fundamental model will further benefit the subsequent models and will be pretty helpful for the improvement of the whole system.

\begin{table}[htbp]
\centering
\caption{The averaged AUC scores of the pre-defined LSTM/BERT architectures and NAS method with different numbers of scenarios.}
\label{Tab:result-initial-model}
\begin{tabular}{c|c|c|c}
\toprule
 Initial Numbers   & LSTM & BERT & NAS \\ \midrule
 2              & 0.731 & 0.733 & \textbf{0.751} \\  \midrule
 4               & 0.749 & 0.748 & \textbf{0.757} \\  \midrule
 8               & 0.762 & 0.761 & \textbf{0.767} \\  \midrule
16               & 0.771 & 0.778 & \textbf{0.783} \\  
\bottomrule
\end{tabular}
\end{table}

\subsubsection{The Benefits of Leveraging Behavior Sequences}
\begin{figure}[hbt]
\centering
 	\includegraphics[width=0.38\textwidth]{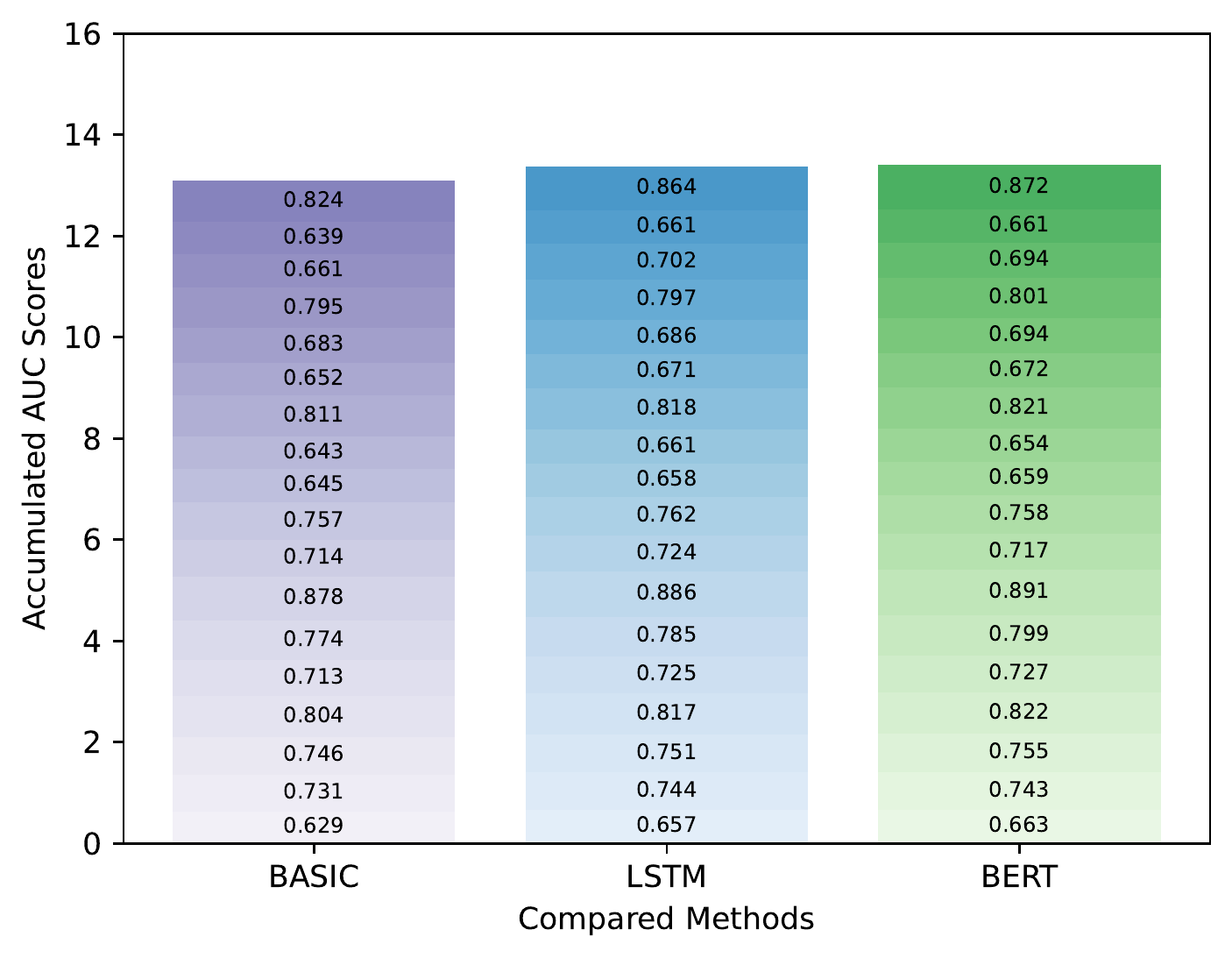}
	\caption{The accumulated AUC scores of the models with and without sequence information.}
	\label{Fig:result-sequences}
\end{figure}

As we discussed before, the use of behavior sequences can always boost the prediction in real-world applications.
To validate that, we also conduct some further experiments with dataset A.
Concretely, we add a basic model with only the profile information used, while the behavior sequences are not considered.
Similar to the SinH strategy, we simply train the model with the data of every single scenario, and the accumulated and averaged results of this model and the LSTM/BERT-based model under SinH strategy are shown in Fig.~\ref{Fig:result-sequences} and Table~\ref{Tab:result-sequences}. 
We can easily conclude from the results that the use of the behavior sequences is obviously helpful in this application, and with a proper design of the model, the performance can be further improved (e.g., the BERT based architecture achieved a $1.70\%$ improvement on average with regard to AUC score).

\begin{table}[ht]
\centering
\caption{The averaged AUC scores with and without sequence information.}
\label{Tab:result-sequences}
\begin{tabular}{c|c|c|c}
\toprule
    & Basic & LSTM & BERT \\ \midrule
AVG & 0.728 &0.743 & \textbf{0.745} \\ \bottomrule
\end{tabular}
\end{table}

\subsubsection{The Results with Different Initial Scenarios}

Note that in most of our previous experiments, we randomly select $8$ scenarios as the initial ones to simulate the setting. 
We further validate the performance when we vary the initial scenarios with dataset A.
Although we can't go through all the different initialization cases, we can get more insight by varying the different initialization scenarios.
Similar to the previous experiment procedure, we vary the number of the initial scenarios to $\{2,4,8,16\}$, and report the associated averaged results of the BERT-based models in Table~\ref{Tab:results-varied-initial}.
Consistent with the previous results in Table~\ref{Tab:result-AUC}, MeH strategy achieved the best performance in all of the situations, while our methods showed a slight disadvantage compared to the MeH strategy.
Moreover, with more initial scenarios, the performance tends to get better. 
We believe that is reasonable since more information is observed and extracted in the initialization stage of the scenario agnostic heavy model, helping it to achieve better generalization across all scenarios. 
In addition, we also conduct experiments to validate the results if we keep the number of initial scenarios as the same (e.g., $8$) while varying the initial selection.
Still, MeH and our strategy show consistent superiority when compared to the other strategies, while the final averaged AUC scores may be changed.
We attribute these changes to the size and representativeness of the initial scenarios.

\begin{table}[htbp]
\centering
\caption{The averaged AUC scores of the compared methods with different numbers of initial scenarios.}
\label{Tab:results-varied-initial}
\begin{tabular}{c|c|c|c|c}
\toprule
 Initial Numbers   & SinH & MeH & MeL & Ours \\ \midrule
 2               & 0.745 & \textbf{0.747} & 0.741 & 0.747 \\  \midrule
 4               & 0.745 & \textbf{0.751} & 0.744 & 0.749 \\  \midrule
 8               & 0.745 & \textbf{0.756} & 0.746 & 0.754 \\  \midrule
16               & 0.745 & \textbf{0.769} & 0.750 & 0.763 \\  
\bottomrule
\end{tabular}
\end{table}

\subsection{Online Application}

To further validate the usefulness of the system, we perform a series of online experiments on a recommendation task, which is associated with $34$ scenarios.
The goal of this task is to provide proper recommendations in each scenario so that a better click-through rate (CTR) can be obtained, which means that the recommendation is more suitable for the users.

\subsubsection{Online Performance}

\begin{figure}[htbp]
\centering
 	\includegraphics[width=0.35\textwidth]{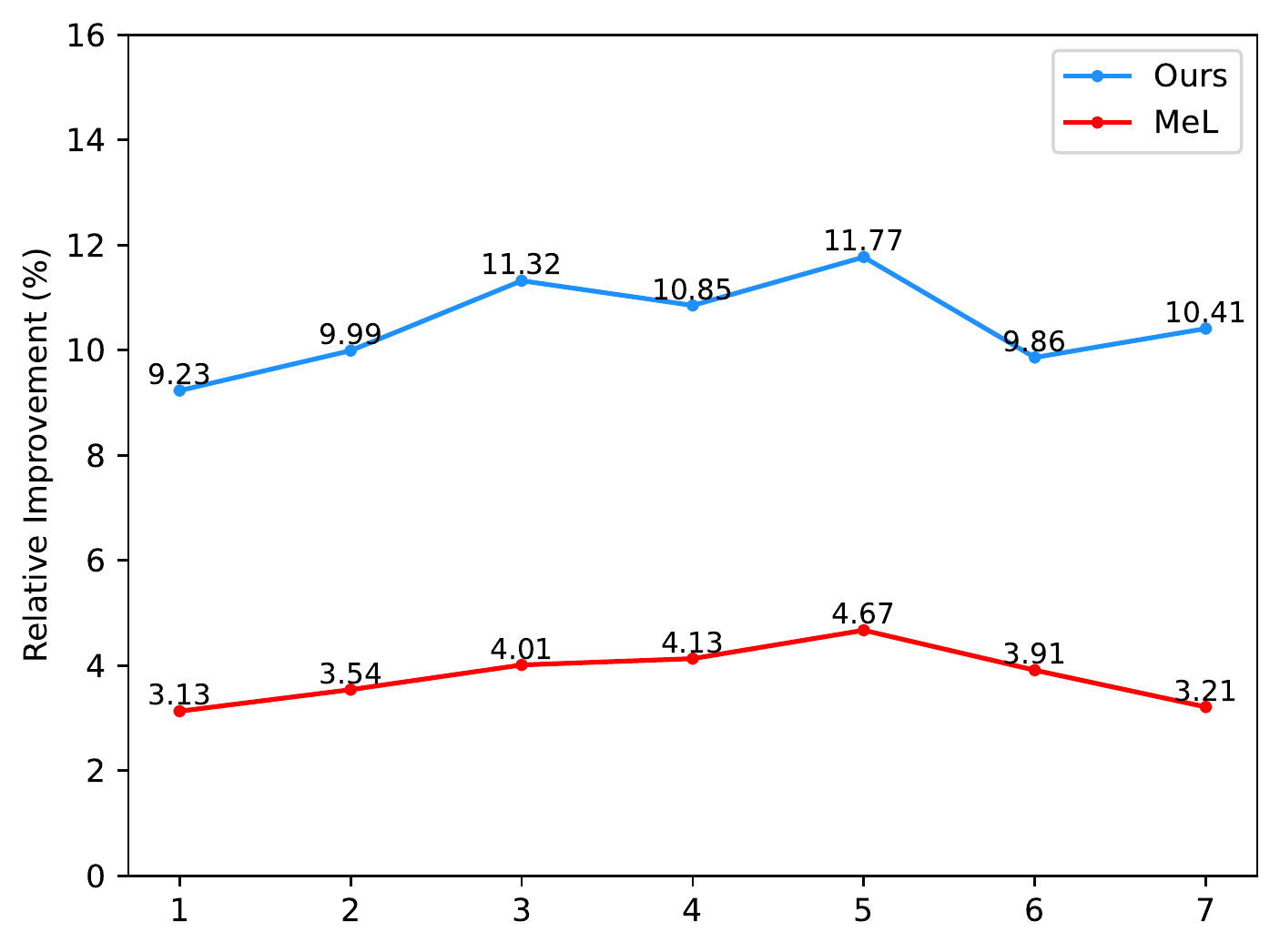}
	\caption{The relative improvement (\%) of the proposed method and MeL strategy compared to the baseline.}
	\label{Fig:Results-Online}
\end{figure}

We compare the models obtained with our proposed system with the models constructed with the specific scenarios, which are similar to the models obtained with the aforementioned SinH strategy, while \textit{lighter} architectures are used in these models to satisfy the efficiency requirement, and necessary efforts are taken to fine-tune the models to boost the performance.
Moreover, we also take the models obtained with the above mentioned MeL strategy into comparison.
The averaged relative improvement results of our method and the MeL strategy with regard to CTR among all of these scenarios are shown in Fig.~\ref{Fig:Results-Online}.
As we can see, in an observational period of $7$ days, our method consistently outperforms the compared strategies, with an averagely $10.49\%$ relative improvement when compared to the basic baseline.
As for the MeL strategy, an averagely $3.80\%$ relative improvement is obtained, compared to the basic baseline.
The superiority of our method over MeL strategy is also obvious in this online application.
These results adequately demonstrate the usefulness of the presented framework.

\subsubsection{Efficiency Discussion}
One issue that needs to be mentioned is that our models are all generated with the developed system, while the compared models are typically fine-tuned by the algorithm experts.
More specifically, to obtain the baseline models, typically a tentative model is first trained and evaluated. After that, some subsequent fine-tuning processes are followed. This will lead to an average time overhead of almost three hours for each of these scenarios.
As for the models generated with the system, aside from the necessary results checks, there is little extra labor cost. The average time cost for each scenario is less than half an hour.
That is to say, our models are efficiently generated without redundant human efforts. And that is really one critical issue required in the similar tasks as we discussed before.
\section{Related Works}
\label{related works}

As an underappreciated problem, inadequate efforts have been taken towards the long tail scenario modeling problem with budget limitation, whilst some other topics may be considered closely related to it, which also provide helpful insights for solving this problem.
Following the philosophy of knowledge sharing, multi-domain modeling~\cite{DBLP:conf/cikm/ShengZZDDLYLZDZ21,DBLP:journals/ml/DredzeKC10,DBLP:conf/cikm/LiLDZZ20} tries to construct models with the data of multiple related domains, and boost the performance of each domain with the knowledge from other domains. However, the emerging scenarios are not addressed in this problem.
The studies of cold-start problem~\cite{DBLP:conf/kdd/BriandSBMT21,DBLP:conf/sigir/HuanZZZWGGHZM22,DBLP:journals/tkde/ZhaoLHCWL16} address the arising of new scenarios with no samples, and aim at effectively outputting predictions or estimations for the emerging scenarios, by leveraging the information of previously related scenarios.
Meta learning~\cite{DBLP:conf/nips/AndrychowiczDCH16,DBLP:conf/icml/FinnAL17,DBLP:conf/iclr/RenTRSSTLZ18,DBLP:conf/icml/SantoroBBWL16,DBLP:journals/csur/WangYKN20,DBLP:conf/nips/PatacchiolaTCOS20,DBLP:conf/kdd/LeeIJCC19,DBLP:conf/sigir/PanLATH19,DBLP:conf/sigir/ZhuXZGSZLC21,zhang2022meta}, as another representative paradigm that utilizes the information of related scenarios to boost the modeling of each specific scenario, can be naturally applied to solve the cold-start problem~\cite{DBLP:conf/kdd/LeeIJCC19,DBLP:conf/sigir/PanLATH19,DBLP:conf/sigir/ZhuXZGSZLC21}. 
And also, meta learning can efficiently build a specific model for new scenarios with limited samples, making it a potential choice while handling long tail scenario modeling.
However, from a more realistic view, the tasks in industrial settings are with not only continually emerging scenarios but also a budget limitation for computing and human resources.
That is, insufficient human efforts can be occupied to deal with each of these scenarios, and limited time and computing resources can be available for the online inference stage.
This situation raises the demand for building an automatic system to cope with the long tail scenario modeling problem, and taking the budget limitation into consideration at the same time.

As a powerful assistant, automatic machine learning related techniques~\cite{hutter2019automated,DBLP:conf/icde/ShiZLYLZ20,DBLP:journals/corr/abs-1810-13306,DBLP:journals/jair/ZollerH21,DBLP:conf/nips/BergstraBBK11,DBLP:conf/icml/PhamGZLD18,DBLP:conf/icml/FinnAL17} have been widely employed among various applications, especially for those in which automatic pipeline is equipped.
Among these techniques, hyperparameter optimization~\cite{DBLP:conf/nips/BergstraBBK11,DBLP:conf/iclr/HazanKY18,DBLP:conf/icml/FalknerKH18} may be the most frequently used, which can be beneficial for the optimization of almost all models. 
Different methods, such as random search~\cite{DBLP:journals/jmlr/BergstraB12}, evolutionary algorithm~\cite{DBLP:journals/ec/HansenMK03}, Bayesian optimization~\cite{DBLP:conf/nips/SnoekLA12}, and classification-based strategy~\cite{DBLP:conf/aaai/YuQH16}, etc., have been proposed during past few decades, and various systems have also been developed~\cite{DBLP:conf/kdd/ThorntonHHL13,komer2014hyperopt,DBLP:conf/kdd/GolovinSMKKS17,H2OAutoML20}.
Neural architecture search (NAS)~\cite{DBLP:conf/iclr/LiuSY19,
DBLP:conf/nips/ChangZGMXP19,DBLP:conf/icml/PhamGZLD18,DBLP:conf/nips/Luo0WQCL20,DBLP:conf/pakdd/RenLYZ22}, the processing of automatically finding a suitable network architecture, has drawn increasing attention during the past few years, and remarkable achievements have been obtained, especially for the tasks that are closely related to computer vision and natural language processing.
Besides, meta learning~\cite{DBLP:conf/nips/AndrychowiczDCH16,DBLP:conf/icml/FinnAL17,DBLP:conf/iclr/RenTRSSTLZ18,DBLP:conf/icml/SantoroBBWL16,DBLP:journals/csur/WangYKN20,DBLP:conf/nips/PatacchiolaTCOS20,DBLP:conf/kdd/LeeIJCC19,DBLP:conf/sigir/PanLATH19,DBLP:conf/sigir/ZhuXZGSZLC21}, also known as learning to learn, is another appealing topic among automatic machine learning communities.
Thanks to its philosophy of leveraging the knowledge of related scenarios, various tasks, such as few-shot learning problem~\cite{DBLP:conf/iclr/RenTRSSTLZ18,DBLP:journals/csur/WangYKN20,DBLP:conf/nips/PatacchiolaTCOS20} and cold-start problem~\cite{DBLP:conf/kdd/LeeIJCC19,DBLP:conf/sigir/PanLATH19,DBLP:conf/sigir/ZhuXZGSZLC21}, have been benefited by using meta learning based techniques.
Targeting the long tail scenario modeling problem, our proposed system gets these helpful tools involved and brings ingenious improvements to establish the whole system.

\section{Conclusion}
\label{conclusion}
In this paper, the problem of long tail scenario modeling with budget limitations is emphasized.
We present an automatic system ALT for handling this problem.
We take several efforts to improve the algorithms used in our system, by employing and proposing various automatic machine learning related techniques.
Moreover, several efforts are also taken to build the whole system.
We further present abundant experimental results and analyses to demonstrate the effectiveness of the proposed system.

\newpage

\bibliographystyle{IEEEtran}
\bibliography{9-icde2023-ref-short} 

\end{document}